\documentclass[conference]{IEEEtran}
\IEEEoverridecommandlockouts
\usepackage{cite}
\usepackage{amsmath,amssymb,amsfonts}
\usepackage{algorithmic}
\usepackage{siunitx}
\usepackage{graphicx}
\usepackage{textcomp}
\usepackage{xcolor}
\usepackage{nicematrix, booktabs}
\usepackage{multirow}
\usepackage{makecell}
\usepackage{enumitem}
\usepackage[a4paper, total={184mm,239mm}]{geometry}
\def\BibTeX{{\rm B\kern-.05em{\sc i\kern-.025em b}\kern-.08em
T\kern-.1667em\lower.7ex\hbox{E}\kern-.125emX}}
\begin{document}

\title{Energy-Efficient \& Real-Time Computer Vision with Intelligent Skipping via Reconfigurable CMOS Image Sensors}

\author{Md Abdullah-Al Kaiser$^{1,\star}$\thanks{$^{\star,\dagger}$Equally contributing authors.} \ \ Sreetama Sarkar$^{2,\star}$ \ \ 
Peter A. Beerel$^{2,\dagger}$ \ \ 
Akhilesh R. Jaiswal$^{1,\dagger}$ \ \
Gourav Datta$^{2}$ \\
$^{1}$University of Wisconsin-Madison, Madison, USA \ \ \ \
$^{2}$University of Southern California, Los Angeles, USA \\
{\tt\small {\{sreetama,pabeerel,gdatta\}@usc.edu} \ \  \tt\small {\{mkaiser8,akhilesh.jaiswal\}}@wisc.edu} \\ \vspace{-6mm}}


\maketitle

\begin{abstract}
Current video-based computer vision (CV) applications typically suffer from high energy consumption due to reading and processing all pixels in a frame, regardless of their significance. While previous works have attempted to reduce this energy by skipping input patches or pixels and using feedback from the end task to guide the skipping algorithm, the skipping is not performed during the sensor read phase. As a result, these methods can not optimize the front-end sensor energy. Moreover, they may not be suitable for real-time applications due to the long latency of modern CV networks that are deployed in the back-end. To address this challenge, this paper presents a custom-designed reconfigurable CMOS image sensor (CIS) system that improves energy efficiency by selectively skipping uneventful regions or rows within a frame during the sensor's readout phase, and the subsequent analog-to-digital conversion (ADC) phase. A novel masking algorithm intelligently directs the skipping process in real-time, optimizing both the front-end sensor and back-end neural networks for applications including autonomous driving and augmented/virtual reality (AR/VR). Our system can also operate in standard mode without skipping, depending on application needs. We evaluate our hardware-algorithm co-design framework on object detection based on BDD100K and ImageNetVID, and gaze estimation based on OpenEDS, achieving up to 53\% reduction in front-end sensor energy while maintaining state-of-the-art (SOTA) accuracy.
\end{abstract}


\section{Introduction}
As the demand for edge computing in computer vision (CV) applications grows—ranging from surveillance systems \cite{surveillance} and autonomous vehicles \cite{auto_driving} to smart devices—energy efficiency has emerged as a key challenge.
A major contributor to energy inefficiency in edge devices is the repetitive reading and processing of redundant frames and pixels, particularly those that do not provide useful information for the task at hand. This problem is further exacerbated in video applications that are processed at high frame rates \cite{9941155}, where large amounts of data are handled unnecessarily. 
Consequently, optimizing the input data processing at the sensor level becomes essential to enable complex CV applications efficiently at the edge.

Previous studies have made strides in addressing temporal redundancy in video processing by skipping frames or selectively processing regions with significant changes \cite{parger2022deltacnn, parger2023motiondeltacnn, Dutson_ICCV_2023, feng2022edgaze}. However, these approaches still require reading entire frames to decide which pixels or patches to process. This implies that pixel readouts are still performed for all regions, leading to continued energy consumption during the analog-to-digital (ADC) conversion process, which remains a bottleneck for energy efficiency. 
Moreover, current methods often rely on the output of the previous frame’s task to predict the significance of pixels or patches \cite{feng2022edgaze, reidy2024hirise, sarkar2024maskvd}, introducing delays in processing the current frame. For instance, while the HiRISE system \cite{reidy2024hirise} compresses the high-resolution images, thereby reducing the energy incurred in processing them, it depends on feedback from downstream tasks to generate high-resolution regions for object detection. This dependency can impede real-time effectiveness and adds complexity due to the back-and-forth data transfer, which is challenging in energy-limited environments. Moreover, methods that skip large portions of pixels often suffer from performance degradation \cite{liSpatiotemporalGated2021, mao2021patchnetshortrangetemplate}, which is problematic for safety-critical applications. 

This underscores the need for an intelligent, lightweight pixel masking algorithm capable of generating real-time region-of-interest (RoI) masks without relying on feedback from downstream tasks. In addition, hardware reconfigurability is crucial to enable the skipping of regions or pixels during the pixel readout phase of the sensor, significantly improving energy efficiency. Towards this end, we propose a low-cost vision transformer (ViT)-based intelligent mask generator network that operates independently of task feedback, ensuring compatibility with real-time constraints and optimizing both accuracy and energy use. We also integrate hardware reconfigurability into existing conventional image sensor systems with minimal overhead, including additional memory banks for binary mask storage and power-gating switches. This helps reduce the energy consumed during the sensor readout phase guided by the mask generator. 
Additionally, the masking algorithm boosts the compute efficiency of the back-end neural network processing the CV application by focusing on the salient regions of interest (RoI) identified by the mask. Note that previous works have explored compressing input data at the sensor level through in-sensor computing approaches, where the initial layers of the computer vision (CV) network are processed directly within the sensor \cite{aps_p2m,aps_p2m1,9939582}. While these approaches offer significant energy savings, they complement our method and can be integrated alongside it to further enhance energy efficiency.

The key contributions of our work are as follows. \\
1) \textbf{Real-time Pixel Masking Algorithm:} We propose a novel lightweight pixel masking algorithm tailored for CMOS image sensors that can create a binary mask in real time based on the significance of the pixels in a scene, without depending on feedback from end tasks. \\
2) \textbf{Reconfigurable Peripheral Hardware Integration:} We develop a reconfigurable sensor hardware system that integrates with the masking algorithm, enabling various sensing operation modes, including standard and two skip modes (row-wise skip and region-wise skip). This system yields the front-end energy savings by allowing pixel readouts to be skipped, guided by the binary mask during the sensor's read phase. \\
3) \textbf{Significant Front-end Energy Savings:} Our proposed algorithm-hardware co-design framework achieves notable improvements in front-end (constituting the sensor and the pixel mask generator) energy efficiency—46\% (row-wise skip), 53\% (row-wise skip), and 52\% (region-wise skip), respectively—while maintaining state-of-the-art accuracy in autonomous driving and AR/VR applications. 

\section{Proposed Method}

\subsection{Pixel Masking Algorithm}

The most commonly used technique for skipping redundant computations in a video sequence is by computing pixel-wise differences between input frames and intermediate feature maps \cite{parger2022deltacnn, Dutson_ICCV_2023}, followed by processing only pixels or patches that exceed a pre-determined threshold. However, this requires reading the current frame and hence does not save sensor read energy. Previous works have proposed predicting RoI, and processing the CV network on the RoI \cite{wang2017fastrcnn, ren2016fasterrcnn, maskrcnn, feng2022edgaze} to save overall computation. However, these methods are either expensive \cite{wang2017fastrcnn, ren2016fasterrcnn, maskrcnn}, or depend on the output of the previous frames \cite{feng2022edgaze, sarkar2024maskvd}. This implies that reading subsequent frames in the sensor side has to be stalled till the output information is obtained, and may hinder real-time deployment. Our goal is to design a lightweight network that can be implemented near the sensor and can predict regions of importance without relying on the output of the back-end model. We propose a transformer-based low-cost region mask generator network that can predict the significance of input patches based on attention scores at the input frame rate without consuming significant energy. 

\begin{figure}[!t]
\centering
\includegraphics[width=0.9\linewidth]{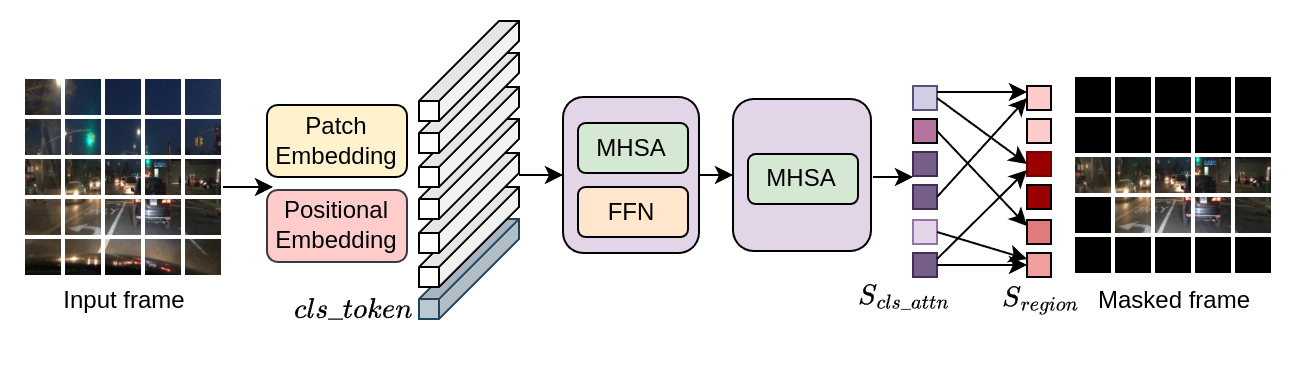}
\vspace{-5mm}
\caption{Mask prediction from input using MGN. MGN consists of a transformer block (MHSA+FFN) followed by a self-attention (MHSA) and linear layer.}
\vspace{-3mm}
\label{fig:region_mask_generator}
\end{figure}

\noindent\textbf{Mask Generator Network:} Our mask generator network (MGN) is built using a single transformer block \cite{dosovitskiy2021an}, followed by a self-attention layer \cite{NIPS2017_3f5ee243} and a linear layer. Initially, the MGN divides the input image into \si{N} patches of size \si{p}$\times$\si{p} and embeds each patch into a vector of length \si{L}. A classification token ($cls\_token$), which is an embedding vector of the same length, is appended to the \si{N} patch embeddings, resulting in $\si{N}+1$ vectors, as illustrated in Figure \ref{fig:region_mask_generator}. A trainable positional embedding is added to each patch embedding before passing the vectors through the transformer encoder block. This encoder consists of a multi-head self-attention (MHSA) layer followed by a feed-forward network (FFN). The embedding vectors are transformed into Query ($\si{Q}$), Key ($\si{K}$), and Value ($\si{V}$) matrices, each of dimension $\left[\si{N}{+}1, \si{d}\right]$, where $\si{d}=\si{L}/\si{H}$ and $\si{H}$ is the number of heads in the MHSA. Self-attention is then performed by each head, as shown in Equation \ref{eq:attn}. The FFN is a 2-layer multi-layer perceptron (MLP) with GELU activation \cite{hendrycks2023gaussianerrorlinearunits}.

\begin{equation}
    {\si{Attention}}(\si{Q}, \si{K}, \si{V}) = {\rm Softmax}(\frac{\si{Q} \si{K}^T}{\sqrt{d}}) \si{V}.
    \label{eq:attn}
\end{equation}

The $cls\_token$ is crucial in ViT-based image classification \cite{dosovitskiy2021an} as it gathers and aggregates information from different patches of the image at the classification head. This aggregation is guided by the attention scores between the $cls\_token$ and the other image patches. Specifically, the attention score, denoted as $\si{S}_{cls\_attn}$, is computed through the dot product of the query vector derived from the $cls\_token$ ($\si{q}_{class}$) and the key matrix from the other patches.

\vspace{-4mm}
\begin{equation}
    {\si{S}_{cls\_attn}} = \frac{\si{q}_{class} \si{K}^T}{\sqrt{d}}.
    \label{eq:stoken}
\end{equation}

$\si{S}_{cls\_attn}$ inherently captures the importance of each patch. Inspired by \cite{liang2022evit}, we utilize this attention score and feed it into a linear layer of the same output dimension as the number image patches to generate the importance scores $\si{S}_{region}$ for the patches. These scores are then passed through a $Sigmoid$ function and thresholded using a region threshold $t_{reg}$ to produce the {Region Mask} ($Mask_{reg}$). This mask contains binary values for each \si{p}$\times$\si{p} region, with \textit{ones} indicating regions to be processed and \textit{zeros} indicating regions to be skipped. Subsequently, the {row mask} ($Mask_{row}$) is derived from $Mask_{reg}$ using a row threshold $t_{row}$. A row is processed if the percentage of active regions in that row exceeds $t_{row}$.

We compute the region mask every \si{P} frames, where \si{P} refers to the interval after which a full-frame is read and a new region mask is predicted. The \textit{identical} predicted mask is used to skip readout and computation in the masked regions for the subsequent $(\si{P}{-}1)$ frames.

\noindent\textbf{Model Training:} The MGN is trained to predict region importance based on object locations within the frame. The ground truth labels are generated from the ground truth bounding boxes or segmentation maps, constructing a 2D binary matrix where regions containing the object, either fully or partially, are marked with \textit{ones}, and regions with no part of the object are marked with \textit{zeros}. The network is trained using binary cross-entropy loss between the predicted region importance scores and the ground truth labels, with performance evaluated through the mean Intersection-over-Union (mIoU) metric.

\subsection{Reconfigurable Hardware Implementation}

To achieve energy savings from sensing and ADC operations for the skipped rows and regions, as identified by our MGN, we design a reconfigurable CMOS image sensor system. This system supports three modes: standard, row-wise skip, and region-wise skip. In the standard mode, the sensor functions like a conventional system, reading each pixel row-by-row with the column-parallel ADC and transmitting the digital bit-streams off-chip for further processing. In row-wise skip mode, the system can bypass reading an entire row in a frame, while in region-wise (patch-wise) skip mode, it skips reading pixels within areas deemed insignificant by the MGN.

\begin{figure}[!t]
\centering
{\includegraphics[width=1\linewidth]{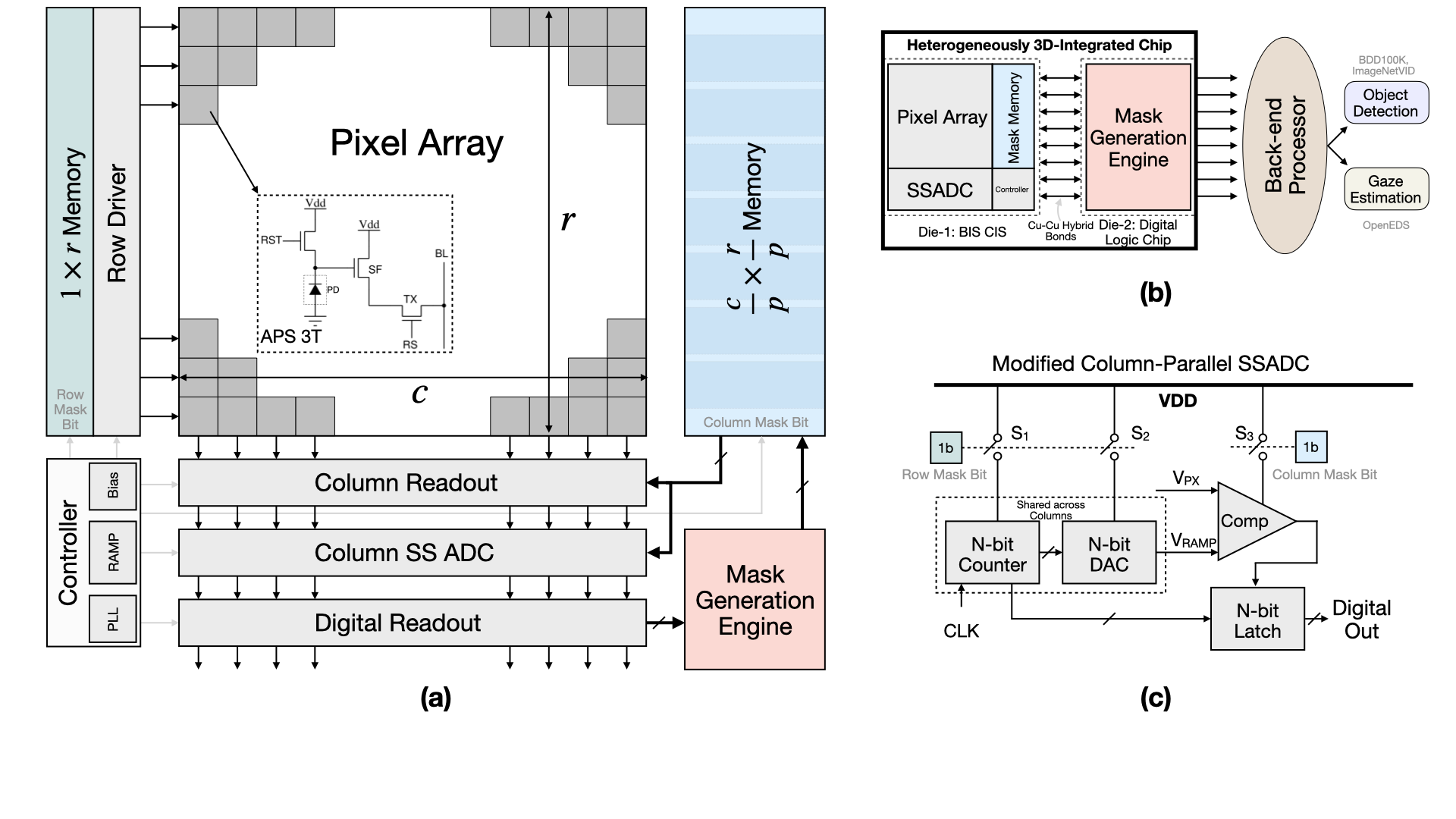}}
\vspace{-10mm}
\caption{(a) Reconfigurable CIS system architecture with mask generation engine and memory blocks, (b) Representative system diagram showing 3D integration of BI-CIS and digital logic chip, and (c) Modified single-slope ADC (SSADC) with power gating for energy-efficient operation.}
\vspace{-5mm}
\label{fig_hardware}
\end{figure}

The proposed reconfigurable CMOS image sensor system, illustrated in Fig. \ref{fig_hardware}(a), includes conventional components, including a pixel array (e.g., 3T or 4T), row driver, ramp generator, counter, column-parallel single-slope ADC (SSADC), output latch, bias circuits, and control circuits \cite{cis_ref1, cis_ref2, cis_ref3}. In addition to these standard blocks, our design incorporates an engine to implement the MGN (that provides the mask information to the sensor), which is illustrated in section II-A. In row-wise skip mode, the engine produces a 1D binary mask that marks significant rows with a value of 1 and insignificant rows with 0. Similarly, in region-wise skip mode, a 2D mask is generated based on pixel importance within the scene. This mask is applied to skip uneventful rows and regions for a number of frames in the row-wise skip and region-wise skip modes, respectively. Hence, our system embeds a 1D memory array of size \si{1{\times}r}, that is connected to the row scanner for the row-wise mode, and a 2D memory array of size \si{\frac{c}{p}{\times}\frac{r}{p}} for the region-wise mode, to store the respective row or region masks. Note that $r$ and $c$ denote the number of rows and columns in the pixel array respectively, and $p$ denotes the patch size employed in our MGN for region-wise skip.

\begin{figure*}[!t]
\centering
{\includegraphics[width=0.8\linewidth]{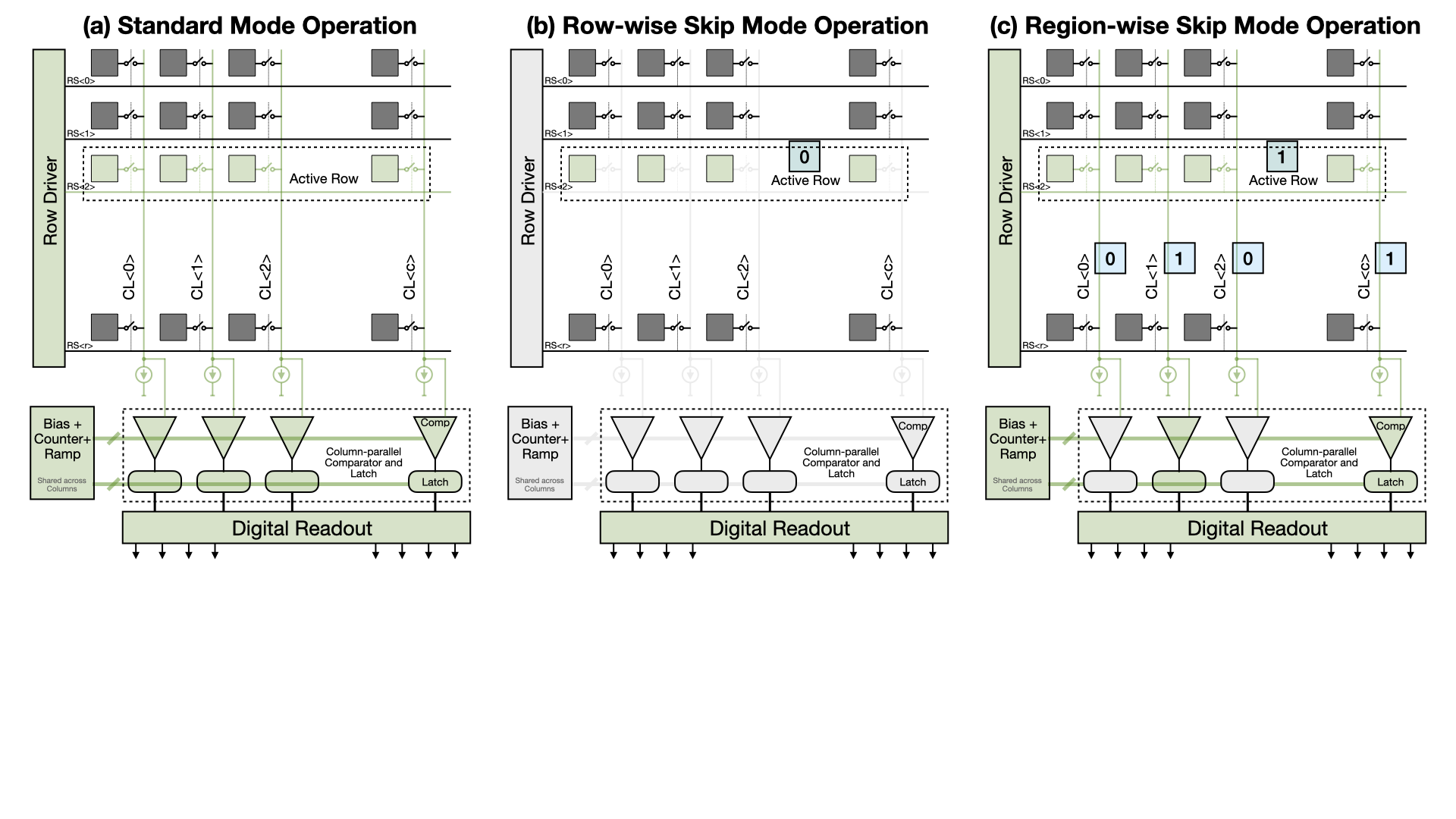}}
\vspace{-30mm}
\caption{Pixel reading operations for (a) standard mode, (b) row-skip mode, and (c) region-skip mode. Active components are highlighted in green, while light gray represents the power-gated elements involved in the reading process across the various modes.}
\vspace{-4mm}
\label{fig_modes}
\end{figure*}

Fig. \ref{fig_hardware}(b) illustrates a representative diagram of the entire system. The back-side illuminated CMOS image sensor (BI-CIS) die, which includes additional memory banks to store the binary mask, is heterogeneously 3D-integrated with a digital logic chip \cite{3D_integ1, 3D_integ2} that implements our mask generation engine. Unlike the sensor die, which is typically fabricated in a lagging process node (e.g., 45 nm), this chip can be manufactured using an advanced process node (e.g., 7 nm), further minimizing the energy overhead of the MGN. Based on the pixel array dimension and patch size employed by the MGN, the total memory requirement to store the binary row and region masks is less than 5 kB, which can either be allocated within the BI-CIS die without significant overhead or can also be integrated into the digital logic chip. Digital bit-streams from the CIS array, along with the generated binary masks from the MGN engine, are communicated between the sensor and logic chips through fine-pitched Cu-Cu hybrid bonding \cite{cu_cu_pitch}. Finally, the digital output generated by the image sensor is transmitted off-chip to a back-end processor for the downstream CV processing, depending on the selected mode (standard, row-wise skip, or region-wise skip).

Fig. \ref{fig_hardware}(c) depicts the modified single-slope ADC (SSADC) used in our design. The ADC consists of a counter, current-steering DAC-based ramp signal generator, comparator, and latch to convert analog pixel data into digital bit-streams \cite{ssadc1, ssadc2}. Our intelligent masking algorithm helps reduce the number of ADC conversions, thereby saving energy. Since the ramp generation circuit is shared across all column ADCs, when the row mask bit is 0 in the row-skip mode, the ramp generation circuit (counter and DAC) can be power-gated using switches S1 and S2, as shown in Fig. \ref{fig_hardware}(c). Additionally, the ADC comparators and latch associated with the column-parallel ADC can be power-gated using switch S3 (Fig. \ref{fig_hardware}(c)). Clock gating is also employed to further reduce energy consumption, along with resetting the latch in the row-skip mode. However, in region-wise skipping, where only some pixels in a row are read, complete power-gating of the ramp generation, and counters is not possible. Instead, we can save energy by power-gating only the comparators and column latches while keeping the ramp generation circuit and counter active. 
Our ADC is designed to be reconfigurable, allowing for precise adjustments to output bit precision by modifying the ramp slope (voltage per clock cycle). By altering the reference current in the ramp generation circuit (which utilizes a pMOS cascode current mirror-based current DAC) or the reference resistance, we can effectively reconfigure the ramp slope \cite{ssadc1}. In this work, we employ a 10-bit ADC, aligning with the traditional pixel bit-depth of 10 \cite{onsemi:AR0135AT} for our CIS system. However, the precision can be adapted for different applications by adjusting the bias current or resistance and resetting the counter at specific intervals. For instance, to switch from 10-bit to 8-bit precision, the counter can be reset after 256 clock cycles. This capability enhances energy efficiency, depending on application-specific requirements.

Fig. \ref{fig_modes} illustrates the various operational modes of our proposed hardware, which are described below. \\
\textbf{Standard Mode:} Fig. \ref{fig_modes}(a) illustrates the standard mode of operation, where the conventional row-by-row reading method is applied. One row is activated at a time, and the analog pixel values are read by the column-parallel ADC, following the conventional procedure. In this mode, the row-mask and column-mask memories are both constantly set to 1. \\ 
\textbf{Row-skip Mode:} In the row-skip mode, entire rows are disabled during the read process. For example, when the row mask bit is 0, indicating an insignificant row, the row driver is power-gated, preventing the activation of the pixel transmission gates (e.g., \si{RS<2>} in Fig. \ref{fig_modes}(b)). Consequently, all bitlines are discharged by the tail current source of the source follower, resulting in no bitline activity, which conserves sensor energy. Additionally, the counter and ramp generation circuits, along with all column-parallel ADCs, are power-gated. The latches are reset, and all 0s are transmitted for the skipped row. For significant rows, where the row mask bit is 1, the conventional reading method is used. \\
\textbf{Region-skip Mode:} In region-skip mode, patch-wise masks for each region in the CMOS image sensor array are stored in both the column and row mask memories. During this process, the row and column masks operate concurrently. If the row mask is 0, the row remains inactive, and the column readout circuits and SSADC are power-gated to conserve the maximum energy. Conversely, if the row mask is 1, indicating that there are active pixels in the row, the column mask memory determines which column-parallel SSADCs will be active. The ADCs associated with the insignificant pixels (where the column mask is 0) will be power-gated, while significant ones (where the column mask is 1) will be read. For example, in Fig. \ref{fig_modes}(c), SSADCs associated with \si{CL<1>}, \si{CL<c>} are actively read, while \si{CL<0>}, \si{CL<2>} are skipped in this region-wise skip mode.

Note that in row-skip and region-skip modes, uneventful pixels are not read, resulting in an N-bit (ADC bit precision) value of $0$ being transmitted for those pixels. This method preserves the same transmission protocol as traditional systems, however, asynchronous reading \cite{aer} can be employed to enhance latency and transmission bandwidth, further reducing the energy.

\section{Experimental Results}
\subsection{Models and Datasets}

\noindent\textbf{Object Detection:} We evaluate our approach on two video object detection datasets: ImageNetVID \cite{russakovsky2015ImageNet} and BDD100K \cite{yu2020bdd100k}. We demonstrate results using ViTDet \cite{Li2022vitdet} on ImageNetVID validation set, which consists of 639 video sequences, spanning 30 categories of objects, with each sequence containing up to 2895 frames. For BDD100K, we report results using single-head Tiny-YOLO \cite{tinyyolo} on 200 test samples, each with a sequence length of 200. The annotated dataset is at 5 FPS whereas the videos in BDD100K are captured at 30 FPS. We perform evaluation with \si{P}=4 at 5 FPS, which translates to \si{P}=24 at 30 FPS. 
Tiny-YOLO on BDD100K is fine-tuned with the input mask for 20 epochs. For ViTDet, no additional fine-tuning is required. Our masking method is applied directly during inference, as described in \cite{sarkar2024maskvd}, to reduce computational overhead in the ViT back-end by processing fewer tokens. 
To prevent accuracy loss, a reference tensor is stored for each transformer block from the last processed frame, ensuring consistent performance across frames while minimizing redundant computations.

\noindent\textbf{Eye Tracking:}
A typical eye-tracking pipeline consists of two stages: (a) feature extraction via eye segmentation and (b) gaze estimation. While gaze estimation is relatively lightweight, eye segmentation constitutes 85\% of the system's overall computational cost \cite{feng2022edgaze}. We focus on the performance bottleneck—eye segmentation—as it largely dictates the accuracy of gaze estimation. Our approach is evaluated using EyeNet \cite{chaudhary2019ritnet}, a U-Net-based eye segmentation network, applied to sequential data from the OpenEDS 2020 \cite{palmero2020openeds2020} dataset. This dataset includes 152 participants with diverse ethnicities and eye colors, captured at 100 FPS with a resolution of 640$\times$400. For our experiments, we use the ground truth segmentation maps generated by \cite{feng2022edgaze}, which cover 185 video sequences totaling 27,431 frames—about 150 frames per 30-second sequence. Since the dataset is annotated at 5 FPS, we incorporate a scaling factor of 20 in the calculation of \si{P} to reflect the actual real-time processing at 100 FPS.

\noindent\textbf{Mask Generator Network:} The MGN consists of a single ViT layer with $\si{p}$=16, $\si{L}$=192, an input resolution 224$\times$224, that results in 1.86M parameters and 0.161 GFLOPs, where FLOPs denotes the number of floating point operations. The initial weights for the mask generator network as well as the $cls\_token$ are restored from a ImageNet pre-trained DeiT-Tiny \cite{touvron2021deit} model. 
The MGN is trained for 20 epochs using Adam optimizer \cite{kingma2014adam} with a learning rate of 0.001. 

\begin{table}[htbp]
\small\addtolength{\tabcolsep}{-4pt}
    \centering
    \caption{Model accuracy vs skip \% with hyperparameter variation for object detection and eye segmentation. The skip \% used for energy estimation in Figure \ref{fig_energy_plot} are highlighted in bold.}
    \begin{tabular}{c|c|c|ccc|c|cc}
    \toprule
      \textbf{Model} & \textbf{Dataset} &	\textbf{Mask} & \textbf{\si{P}}	& \textbf{$t_{reg}$} & \textbf{$t_{row}$}	& \textbf{mAP/} & \textbf{Skip\%} & \textbf{Skip\%} \\
      & & & & & & \textbf{mIoU} & \textbf{(pixel)} &
      \textbf{(row)}  \\ 
      \midrule
      \multirow{7}{*}{\makecell{Tiny-\\YOLO}} & \multirow{7}{*}{\makecell{BDD-\\100K}} & Baseline & - & - & - & 0.237 & 0.0 & 0.0 \\
      \cmidrule{3-9}
      & & Region & 24 & 0.05 & - & 0.236 & 0.45 & 0.26 \\
      & & & 24 & 0.1 & - & \textbf{0.233} & \textbf{0.57} & \textbf{0.38}\\
      & & & 24 & 0.2 & - & 0.218 & 0.69 & 0.51\\
      \cmidrule{3-9}
      & & Row & 24 & 0.05 & 0.5 & 0.239 & 0.45 & 0.45\\
      & & & 24 & 0.1 & 0.5 & \textbf{0.233} & \textbf{0.58} & \textbf{0.58}\\
      & & & 24 & 0.2 & 0.5 & 0.215 & 0.70 & 0.70\\  
            \midrule
      \multirow{3}{*}{ViTDet} & \multirow{3}{*}{\makecell{ImageNet\\VID}} & Baseline & - & - & - & 0.823 & 0.0 & 0.0\\
    \cmidrule{3-9}
      & & Region & 24 & 0.5 & - & \textbf{0.815} & \textbf{0.70} & \textbf{0.41}\\
    \cmidrule{3-9}
      & & Row & 24 & 0.5 & 0.1 & \textbf{0.81} & \textbf{0.65} & \textbf{0.65} \\

      \midrule
      \multirow{7}{*}{EyeNet} & \multirow{7}{*}{\makecell{Open\\EDS}} & Baseline & - & - & - & 0.988 & 0.0 & 0.0 \\
    \cmidrule{3-9}
      & & Region & 80 & 0.05 & - & 0.967 & 0.80 & 0.57\\
      & & & 160 & 0.05 & - & \textbf{0.964} & \textbf{0.80} & \textbf{0.57}\\
      & & & 200 & 0.05 & - & 0.959 & 0.80 & 0.57\\
      \cmidrule{3-9}
      & & Row & 80 & 0.05 & 0.1 & 0.974 & 0.59 & 0.59\\   
      & & & 160 & 0.05 & 0.1 & \textbf{0.972} & \textbf{0.59} & \textbf{0.59}\\
      & & & 160 & 0.05 & 0.3 & 0.942 & 0.67 & 0.67\\
      \bottomrule
    \end{tabular}
    \vspace{-3mm}
    \label{tab:acc_vs_skip}
\end{table}

\subsection{Performance Analysis}

Table \ref{tab:acc_vs_skip} showcases the accuracy vs skip (or energy) ratios achieved using our MGN with variations in hyperparameters applied to two tasks across three datasets. We evaluate object detection using mAP-50 and eye segmentation using mIoU. The key hyperparameters governing the trade-off between accuracy and energy are \si{P}, $t_{reg}$ and $t_{row}$. Increasing $t_{reg}$ and $t_{row}$ leads to skipping a large number of pixels and rows, which enhances energy efficiency but at the expense of mAP/mIoU. On the other hand, increasing \si{P} extends the interval between full-frame reads and mask re-computation, reducing the computational overhead of mask generation per frame but also impacting mAP/mIoU. Our MGN demonstrates the ability to skip up to 57\%, 70\%, and 80\% pixels in masked frames on BDD100K, ImageNetVID, and OpenEDS in region-skip mode with accuracy degradation of 0.4\%, 0.8\%, and 2.4\%, respectively. In row-skip mode, we achieve skip rates of up to 58\%, 65\% and 59\% with accuracy degradation of 0.4\%, 1.3\% and 1.6\% respectively. We choose skip ratios that provide an optimal trade-off and maintain mAP/mIoU close to the baseline ($<$1.5\% for detection and $<$2.5\% for segmentation) for front-end energy estimation (bolded in Table \ref{tab:acc_vs_skip}). Since transformer-based networks process data in a patch-wise manner, the reduction in back-end computation is roughly proportional to the skip ratio of input patches. In CNNs, this compute reduction can be achieved by focusing on the RoI. However, due to the mask shape constraints, additional regions may need to be processed, resulting in energy savings, that are bounded by the skip ratio.

\begin{table}
\small\addtolength{\tabcolsep}{-4pt}
    \centering
        \caption{Comparison of accuracy and backend FLOPs with existing efficient video object detection approaches on ImageNetVID}
    \begin{tabular}{c|c|c|c}
    \toprule
    \textbf{Method} & \textbf{Backbone} & \textbf{mAP-50} & \textbf{GFLOPs } \\
    \midrule
    PatchNet \cite{mao2021patchnetshortrangetemplate} & ResNet-101 & 0.731 & 34.2  \\
    \midrule
    ViTDet \cite{Li2022vitdet} & ViT-B & 0.823 & 174.9 \\
    \midrule
    Eventful \cite{Dutson_ICCV_2023} & ViT-B & 0.822 & 87.9 \\
    \midrule
    MaskVD \cite{sarkar2024maskvd} & ViT-B & 0.819 & 94.98 \\
    \midrule
    Ours (region) & ViT-B & 0.815 & 95.86\\
    Ours (row) & ViT-B & 0.810 & 100.27\\
    \bottomrule
    \end{tabular}
    \vspace{-3mm}
    \label{tab:imagenetvid}
\end{table}

\noindent\textbf{Comparison with exiting approaches:} In Table \ref{tab:imagenetvid}, we compare the mAP-50 and FLOPs of the back-end CV network with our approach against existing efficient video object detection methods on ImageNetVID. While PatchNet \cite{mao2021patchnetshortrangetemplate} achieves the lowest FLOPs, it suffers from significant accuracy degradation. In contrast, our method maintains mAP-50 comparable to state-of-the-art approaches with less than 1\% accuracy drop for region skipping and ${\sim}$1.3\% degradation for row skipping.
Notably, unlike existing methods, which only determine which patches to process after reading the entire frame, our approach skips patches during the readout stage, making the task considerably more challenging but more energy-efficient.

\begin{figure}
    \centering
    \includegraphics[width=0.7\linewidth]{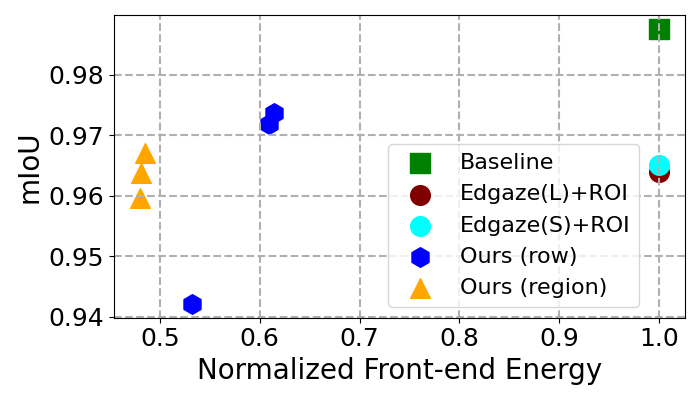}
    \vspace{-1mm}
    \caption{Comparison of mIoU vs normalized front-end energy reduction with existing approaches on OpenEDS dataset}
    \vspace{-4mm}
    \label{fig:eye_tracking_miou}
\end{figure}

Figure \ref{fig:eye_tracking_miou} compares our method against current eye-tracking baselines in terms of both accuracy and front-end energy. We evaluate against EyeNet \cite{chaudhary2019ritnet} and its light-weight variants proposed in Edgaze\cite{feng2022edgaze}. Unlike our approach, Edgaze \cite{feng2022edgaze} performs ROI prediction based on the current frame and can not skip pixels during the readout stage. As a result, we assume Edgaze consumes the same front-end energy as the baseline. Our method matches the accuracy of Edgaze while reducing the front-end energy by 58\%. Additionally, our approach yields similar back-end processing energy savings as Edgaze.

\subsection{Qualitative Analysis}
\begin{figure}[htbp]
  \centering
      \includegraphics[width=0.3\linewidth]{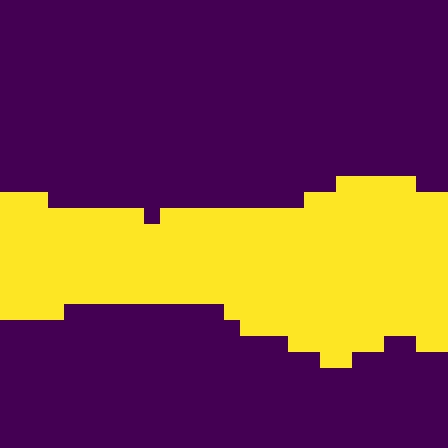}
      \includegraphics[width=0.3\linewidth]{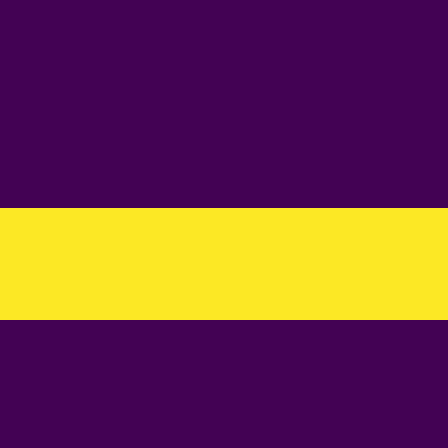}
      \includegraphics[width=0.3\linewidth]{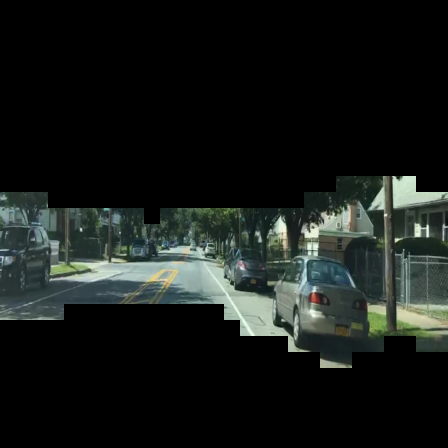} 
    \vspace{-2mm}
      \caption{Visualization of masks on BDD100K showing (a) region mask with $t_{reg}$=0.1, (b) row mask with $t_{row}$=0.5, and (c) input masked with (a).}
    \vspace{-6mm}
      \label{fig:viz_bdd}
\end{figure}

\begin{figure}[htbp]
  \centering
      \includegraphics[width=0.3\linewidth]{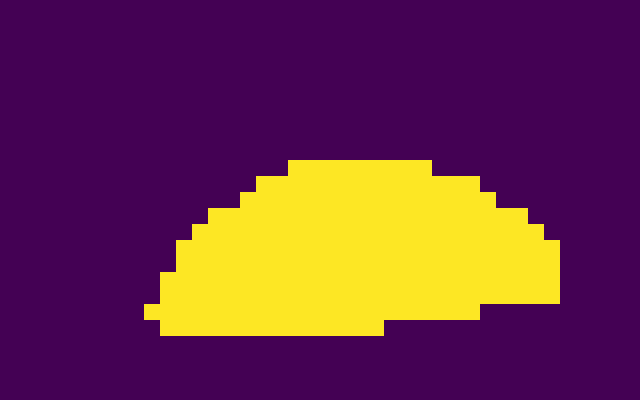}
      \includegraphics[width=0.3\linewidth]{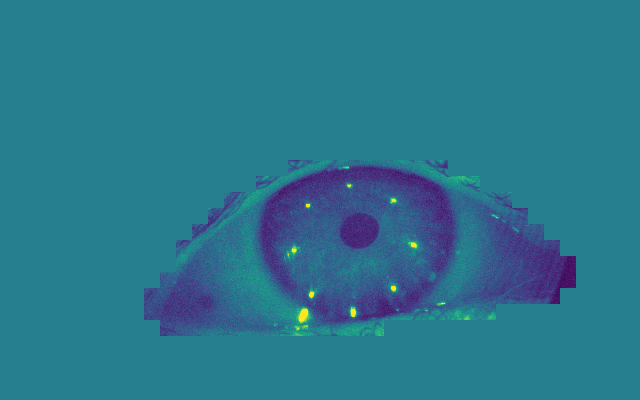}
      \includegraphics[width=0.3\linewidth]{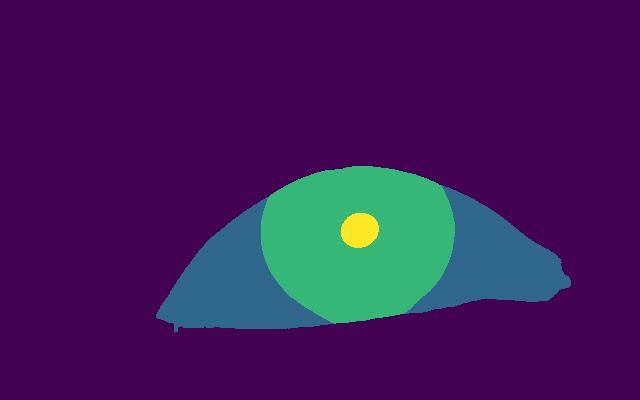} 
      \vspace{-2mm}
      \caption{Visualization of masks on OpenEDS showing (a) region mask with $t_{reg}$=0.05, (b) masked input with (a), and (c) segmentation output.}
      \vspace{-4mm}
      \label{fig:viz_eye}
\end{figure}

The visualization of masks in Figures \ref{fig:viz_bdd} and \ref{fig:viz_eye} demonstrates that MGN accurately captures the important regions at a small fraction of the cost (2\% for EyeNet and 0.09\% for ViTDet) of detection and segmentation networks.

\subsection{Front-end Energy Analysis}

\begin{figure}[!t]
\centering
\vspace{-2mm}
{\includegraphics[width=0.75\linewidth]{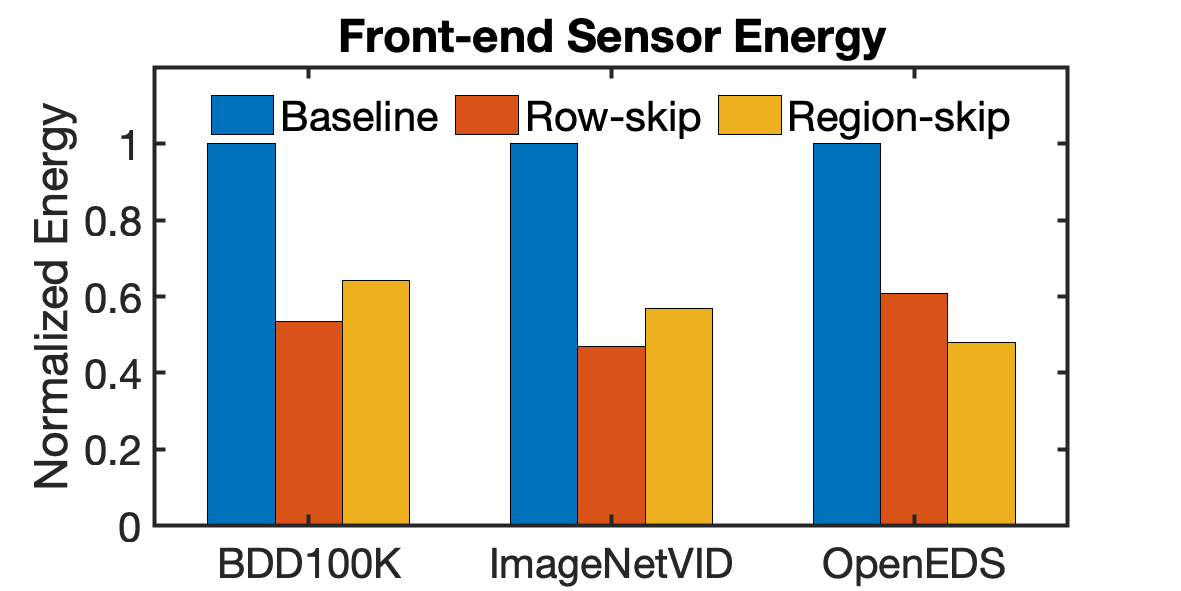}}
\vspace{-2mm}
\caption{Normalized front-end sensor energy plot for the row-skip and region-skip mode compared to the baseline for three datasets (BDD100K, ImageNetVID, and OpenEDS).}
\vspace{-5mm}
\label{fig_energy_plot}
\end{figure}

To determine the front-end energy, we simulate our designed system using the GF22FDX node. We evaluate the front-end energy for the three datasets discussed above, accounting for the additional energy required for memory, data communication through the 3D integrated chip \cite{tsv_energy}, and the MGN engine. In our approach, a full frame is read periodically to compute the mask for skipping less relevant rows or regions in subsequent frames. Consequently, we calculate the average energy for both row-skip and region-skip modes based on the relative energy contributions from the fully read and masked frames, as depicted in Eq. \ref{eq_avg_fenergy}. Our energy estimation technique closely follow \cite{glsvlsi,icassp}.

\vspace{-4mm}
\begin{equation}
    E_{F} = \big( E_{F,base} + E_{mem} + (P-1) \times E_{F,mode} \big)/ P + E_{M}
    \label{eq_avg_fenergy}
\end{equation}

Here, $E_{F,base}$ represents the baseline energy for a standard mode frame read, while $E_{mem}$ accounts for the memory energy required to set all the row and column masks to 1 to enable the standard mode, allowing the system to read every pixel in a frame without skipping. $E_{F,mode}$ denotes the energy for row-skip or region-skip modes, \si{P} denotes period after which a full frame is read, and $E_M$ denotes the MGN energy per frame.

The 1.86M parameters in our MGN necessitates a total memory of 1.86 MB (assuming 8-bit weights) which can be integrated within the logic chip where the MGN is implemented. The energy of the MGN is estimated by accounting for both the compute operations, specifically the multiply-and-accumulate operations, and the memory access energy required to retrieve the parameters. We estimate these using our in-house circuit simulations at 22nm CMOS technology and subsequently scale the results to 7nm using technology scaling methodologies for advanced process nodes \cite{technology_scaling}. The $E_{F,mode}$ is computed as

\vspace{-6mm}
\begin{multline}
E_{F,mode} =  E_{mem} + E_{com} + \big(e_{sense,r} + e_{adc,r} \big) \times n_{px,read}  + \\
                     \hspace{20mm} \big(e_{sense,s} + e_{adc,s} \big) \times n_{px,skip} 
                     \label{eq_mode_fenergy}
\end{multline}

Here, $E_{com}$ denotes to the communication energy between the CMOS BI-CIS and the digital logic die (MGN), while $e_{sense,r}$ and $e_{sense,s}$ denote the pixel sensing energy for conventional and skip-mode read operations, respectively. Similarly, $e_{adc,r}$ and $e_{adc,s}$ represent the ADC energy for conventional and skip-mode reads respectively. $n_{px,read}$ and $n_{px,skip}$ denote the number of pixels with mask value of 1 and 0, respectively. 
Fig. \ref{fig_energy_plot} displays the normalized front-end sensor energy relative to the baseline (standard mode) for both row-skip and region-skip modes. The figure illustrates that row-skip mode exhibits better energy efficiency compared to region-skip mode for the BDD100K and ImageNetVID datasets, whereas region-skip mode performs better for the OpenEDS dataset. 

\subsection{Row-wise vs Region-wise skipping}

\begin{figure}[!t]
\centering
\vspace{-2mm}
{\includegraphics[width=0.75\linewidth]{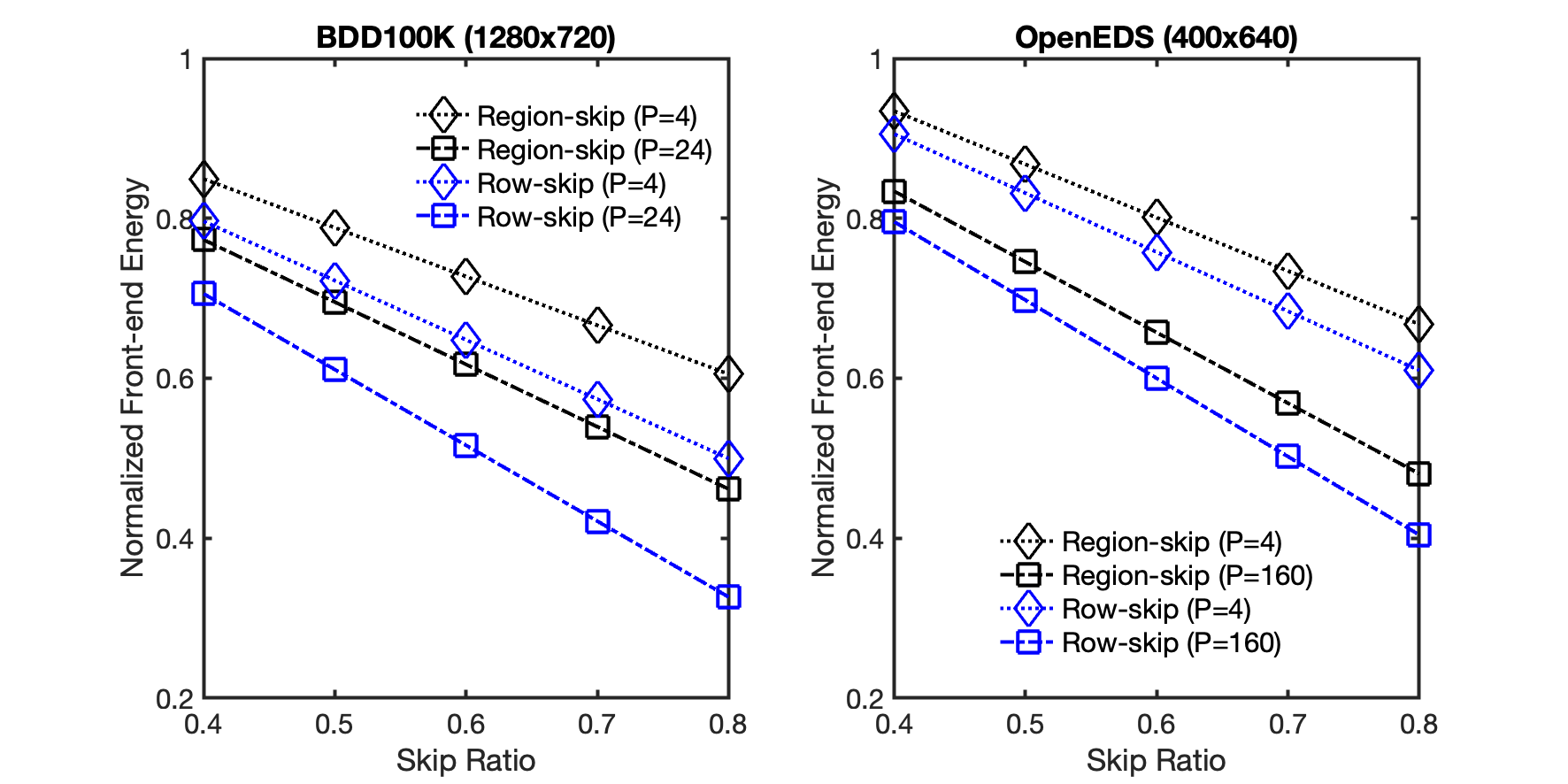}}
\vspace{-2mm}
\caption{Normalized front-end energy estimation plot comparing row-skip and region-skip modes against the skip ratio for the BDD100K (left subplot) and OpenEDS (right subplot) datasets. Here, \si{P} represents the frame period, where one full frame is used to generate the binary masks, followed by masked read operations across the subsequent \si{(P{-}1}) frames.}
\vspace{-4mm}
\label{fig_energy_comp}
\end{figure}

In the standard CIS system, rows are activated sequentially, and all pixels in a row are read in parallel using column-parallel ADCs. The row drivers and part of the SSADCs (including the counter and ramp generator) are shared across columns. In row-skip mode, row drivers and all the ADC blocks (shared and column-parallel) can be power-gated, allowing for skipping entire rows in the frame, thus maximizing energy savings. However, in region-skip mode, row drivers and shared ADC components (counter and ramp generator) must remain active to read the significant pixels in the row. Therefore, only the column-parallel components can be power-gated, leading to lower energy savings compared to row-skip mode for the same skip ratio. Note that by adding a switch between the bitline and the source-follower tail current source, energy efficiency can be further improved for skipped pixels in the region-skip mode. However, this work does not incorporate this extra switch to preserve the traditional CIS system structure. The normalized front-end sensor energy for the two modes for various skip ratios is illustrated in Fig. \ref{fig_energy_comp}. 

The normalized front-end sensor energy improvement is closely linked to the skip ratio. While our MGN aims for state-of-the-art accuracy at higher skip ratios, the energy savings vary between modes depending on these ratios. For example, in the BDD100K dataset, the skip ratio is 0.58 in row-skip mode and 0.57 in region-skip mode. Despite similar skip ratios, row-skip mode achieves greater energy efficiency due to larger energy savings (see the left subplot of Fig. \ref{fig_energy_comp}, with $P{=}$24). Conversely, in the OpenEDS dataset, the skip ratio is 0.59 in row-skip mode and 0.80 in region-skip mode. Here, the higher skip ratio in region-skip mode results in lower energy (right subplot of Fig. \ref{fig_energy_comp}, with $P{=}$160), demonstrating that the increased skip ratio in region-skip mode outweighs the energy benefits of row-skip mode. In addition, energy savings also depend on the total number of masked frames (\si{P{-}1}); higher frame period (P) results in greater energy reductions. Pixel array size also influences energy savings, with larger arrays offering higher energy savings by minimizing the number of ADC reads. Overall, our algorithm-hardware co-design approach is essential to maximize the skip ratio and number of masked frames while maintaining accuracy and reducing front-end energy.

\section{Discussions}

This paper introduces a reconfigurable CMOS image sensor (CIS) system designed to enhance front-end energy efficiency for video-based complex CV tasks by selectively skipping rows or regions during the sensor readout phase, all while maintaining minimal overhead compared to traditional architectures. The system utilizes an intelligent masking algorithm capable of generating real-time RoI masks without relying on feedback from end tasks, making it particularly suitable for real-time applications in energy-constrained environments. We evaluate our system across autonomous driving and AR/VR applications demonstrating significant energy reduction while maintaining high accuracy. Furthermore, our approach can be employed in edge applications utilizing large-format and high-resolution cameras, offering strong potential for low-power, real-time computer vision technologies at high frame rates. By integrating intelligent algorithms with hardware reconfigurability, this method presents significant potential for advancing low-power, real-time computer vision technologies where images are captured at a high frame rate.
\vspace{-1mm}
\section{Acknowledgment}

This work is supported in part by National Science Foundation under award CCF2319617. This work is also supported in part by gift fundings from Samsung and Intel Neuromorphic Research Lab (INRC). In particular, we would like to thank Sumit Bam Shrestha from INRC for fruitful discussions.

{\small
\bibliographystyle{ieee_fullname}
\bibliography{ref}
}

\end{document}